%% file: main.tex
\documentclass[10pt]{article} 
\usepackage[preprint]{tmlr}

\input{math_commands.tex}

\usepackage{amsmath}
\usepackage{amssymb}

\usepackage{url}
\usepackage{xcolor}
\usepackage{hyperref}
\usepackage{graphicx}
\usepackage{subcaption}
\usepackage{comment}

\usepackage{multirow}

\newcommand*{\oldneg}{\mathord{\sim}}
\newcommand\norm[1]{\left\lVert#1\right\rVert}

\newcommand{\vectorize}[1]{\text{vec}\left(#1\right)}

\title{Rank Suggestion in Non-negative Matrix Factorization:
\\Residual Sensitivity to Initial Conditions (RSIC)}

\author{\name Marc A. Tunnell \email tunnellm@mail.gvsu.edu \\
      \addr Grand Valley State University \\
      \AND
      \name Zachary J. DeBruine \email debruinz@gvsu.edu \\
      \addr Grand Valley State University
      \AND
      \name Erin Carrier \email carrieer@gvsu.edu\\
      \addr Grand Valley State University}

\def\month{MM}  
\def\year{YYYY} 
\def\openreview{\url{https://openreview.net/forum?id=XXXX}} 

\newcommand{\doubleblind}[1]{#1}

\begin{document}

\maketitle

\begin{abstract}

Determining the appropriate rank in Non-negative Matrix Factorization (NMF) is a critical challenge that often requires extensive parameter tuning and domain-specific knowledge.
Traditional methods for rank determination focus on identifying a single optimal rank, which may not capture the complex structure inherent in real-world datasets.
In this study, we introduce a novel approach called Residual Sensitivity to Intial Conditions (RSIC) that suggests potentially multiple ranks of interest by analyzing the sensitivity of the relative residuals (e.g. relative reconstruction error) to different initializations.
By computing the Mean Coordinatewise Interquartile Range (MCI) of the residuals across multiple random initializations, our method identifies regions where the NMF solutions are less sensitive to initial conditions and potentially more meaningful.
We evaluate RSIC on a diverse set of datasets, including single-cell gene expression data, image data, and text data, and compare it against current state-of-the-art existing rank determination methods.
Our experiments demonstrate that RSIC effectively identifies relevant ranks consistent with the underlying structure of the data, outperforming traditional methods in scenarios where they are computationally infeasible or less accurate.  
This approach provides a more scalable and generalizable solution for rank determination in NMF that does not rely on domain-specific knowledge or assumptions.
\end{abstract}

\section{Introduction}\label{sec:Introduction}

Low-dimensional models of high-dimensional data are foundational for exploratory data analyses.
Non-negative Matrix Factorization (NMF) has emerged as one such tool for data decomposition and analysis in various domains, including image processing \citep{guillamet_icopr_2002,lee_springer_1999,haifend_tpami_2012}, text mining \citep{hassani_arxiv_2019,pauca_siam_2004}, and bioinformatics \citep{devarajan_plos_2008,gaujoux_springer_2010}.
By decomposing a non-negative matrix into non-negative factors, NMF is often able to extract meaningful patterns and components from complex datasets \citep{gillis_siam_2020}.
However, a critical challenge in applying NMF is determining the appropriate rank of decomposition \citep{wang_tkde_2013}, which essentially dictates the number of components to extract from the data.

Traditionally, rank determination methods in NMF have largely focused on identifying a single ``optimal'' rank using heuristic methods or by leveraging additional knowledge of the distribution of the input data.
They often require arbitrary parameter choices, are sensitive to the initialization, or depend on domain-specific knowledge, which may not always be available or easily interpretable.
While often useful, these methods have considerable limitations.
In this study, we introduce a novel approach to rank determination that seeks to suggest a number of ranks of interest instead of a single optimal rank.

Our method, Residual Sensitivity to Initial Conditions (RSIC), is based on the observation that the reconstruction error of NMF is highly sensitive to the initial conditions of the factorization.
This approach is grounded in a multi-resolution perspective by considering the stability of a rank's residual to its initial conditions.
By doing so, we look to open up new avenues for interpreting NMF results, especially on complex datasets where a single rank may not be able to capture all relevant information.

Our method is designed to be general and applicable to a wide range of datasets, without requiring domain-specific knowledge or assumptions.
Other methods, such as consensus-matrix methods, self-comparison methods, and cross-validation based approaches, have been proposed in the literature to determine the rank of NMF.
The vast majority of rank selection techniques have a strong preference for lower ranks, which may not always be appropriate for the data at hand.
Our method on the other hand, does not have an algorithmic bias for lower ranks and is designed to suggest multiple ranks.
Our methodology is applicable across a wide range of domains and does not rely on domain-specific parameters or \textit{a priori} assumptions about the distribution of the data.

This paper is organized as follows.
Relevant background information on NMF and the methods against which we compare are given in \autoref{sec:Background}.
A detailed description of our method is given in \autoref{sec:Method}.
A high-level description of the datasets we compare on is given in \autoref{sec:Datasets}.
Our experimental setup is given in \autoref{sec:ExperimentalSetup}.
The results are given in \autoref{sec:Results}.
Finally, a discussion and conclusion is given in \autoref{sec:Conclusion}.

\section{Background}\label{sec:Background}
\subsection{Non-negative Matrix Factorization}\label{sec:NMFExplain}
NMF is a low-rank matrix decomposition of an $m \times n$ non-negative matrix, $\mA$, in which non-negativity constraints are imposed in computation of the lower rank factor matrices.
Given a rank $k$ decomposition, the factor matrices $\mW$ and $\mH$ are $m \times k$ and $k \times n$ in dimension, respectively.
Although a variety methods for solving for $\mW$ and $\mH$ exist, such as Hierarchical Alternating Least Squares \citep{kimura_pmlr_2015,gillis_arxiv_2011} or Gradient Descent \citep{lee_nips_2000}, we use Sequential Coordinate Descent (SCD) \citep{franc_springer_2005,lin_neural_2007,hsieh_sigkdd_2011} and Multiplicative Update (MU) \citep{lee_nips_2000,lin_neural_2007} exclusively in this study .
Additionally, although there exists a variety of objective functions such as Kullback-Leibler divergence \citep{lee_nips_2000} and Itakura-Saito divergence \citep{fevotte_neural_2009}, to allow for easier comparison with other work in this field, we minimize the Euclidean distance between $\mA$ and the reconstruction, formulated as
\begin{equation}
    \frac{1}{2}\norm{\mA - \mW \mH}_F^2.\label{eq:Euclidean}
\end{equation}
Given this minimization problem, which is subject to non-negativity constraints, and letting $\mB_W = \mA^T \mW$, $\mB_H = \mA \mH^T$, $\mG_W$ be the gram matrix of $\mW$, and $\mG_H$ be the gram matrix of $\mH^T$, the SCD update rules can be written in vector form as
\begin{align*}
    \mH_{i, :} &\longleftarrow \max\left(0,\mH_{i, :} + \frac{(\mB_W)_{i, :} - (\mH^T \mG_W)_{i, :}}{(\mG_W)_{i, i}}\right),\\
\intertext{ for all $i \in \{1, 2, \dots, k\}$, and}
    \mW_{:, j} &\longleftarrow \max\left(0,\mW_{:, j} + \frac{(\mB_H)_{:, j} - (\mW \mG_H)_{:, j}}{(\mG_H)_{j, j}}\right),
\end{align*}
for all $j \in \{1, 2, \dots, k\}$ \citep{lin_neural_2007,hsieh_sigkdd_2011,franc_springer_2005}.
Note that the max function is applied element-wise.
Similarly, the MU rules can be written as
\begin{align*}
    \mH &\longleftarrow \mH \odot \left(\mB_W^T \oslash \left(\mG_W \mH\right)\right)\\
    \mW &\longleftarrow \mW \odot \left(\mB_H \oslash \left(\mW \mG_H\right)\right),
\end{align*}
where $\odot$ and $\oslash$ denote the Hadamard product and division, respectively \citep{lee_nips_2000,lin_neural_2007}.

Under the view of NMF as an algorithm which produces a clustering, the matrix $\mW$ assigns weights to basis vectors in $\mH$, indicating how strongly each data point is associated with different clusters.
These clusters, which are represented by the basis vectors of $\mH$, often group similar data points together in a meaningful way.

\subsection{Limitations of Non-negative Matrix Factorizations}\label{sec:Limitation}
While NMF has been shown to be useful due to its ability to extract meaningful information, it has shortcomings.
First, the NMF factorization is not unique, making the problem of computing the NMF ill-posed \citep{gillis_jmlr_2012}.
There are typically numerous equally as good solutions (e.g. factorizations for which the Frobenius norm of the residual are equally small).
Second, it is well understood that NMF is highly sensitive to the initial conditions of the problem, and considerable work has gone into getting around this problem, though it largely domain-dependent
\citep{rosales_bio_2016,yang_springer_2021,devarajan_plos_2008}
Third, the underlying optimization problem is generally non-convex, meaning there is no guarantee that the obtained solution is a global minimum (e.g. there is no guarantee that the obtained factorization is the best factorization).

Fundamentally, NMF is a low-rank decomposition, where the factorization can often be meaningfully interpreted as clustering.
However, without additional constraints or sparsity enforcing conditions, the NMF is not a hard clustering algorithm \citep{kim_2008}.
Whether used for a soft clustering or not, in order for the factorization to be useful, it is crucial to know which rank decompositions are meaningful.
Significant work has been done to this effect (as discussed in \autoref{sec:PreviousWork}, with most studies focusing on classification datasets (for which class labels are known).
Typically, this is to ensure that there is a ``true rank''.
While we adopt this convention for consistency and for lack of any better evaluation to date, it is crucial to recognize that NMF is not a classification algorithm (e.g. it is unsupervised and has no knowledge of the underlying classes).
While it is often used for clustering (and it is not uncommon to evaluate clustering with classification data), fundamentally the decomposition of basis vectors, even at the same rank as the number of classes, may have no correlation with the classes considered as the true classes and may be meaningless.

\subsection{Previous Work on Rank Determination}\label{sec:PreviousWork}
A variety of methods to determine the rank have been proposed in the literature.
These can largely be broken into three main categories: consensus-matrix methods, self-comparison methods, and cross-validation based approaches.
A brief overview for each of these methods is provided in this section and further details for the implementations we use for all methods we compare against are given in \autoref{sec:ExperimentalSetup} .

\subsubsection{Consensus-Matrix Methods}\label{subsub:Consensus}
Cophenetic correlation and dispersion coefficient both compute a consensus matrix based on the clustering obtained with NMF, then compute their metric based on this matrix \citep{kim_oup_2007,brunet_nas_2004}.
In both cases, the consensus matrix is computed and then averaged over a number of random starts.

After computing the averaged consensus matrix, both methods compute their metric.
Cophenetic correlation then chooses a rank based on when the cophenetic correlation first begins to drop \citep{brunet_nas_2004}.  
However, it is important to note that what constitutes a drop is dependent on how many ranks are plotted as this affects the range of the $y$-axis and consequently what appears to be a drop.
Alternatively, the dispersion coefficient method chooses a rank based on when the dispersion coefficient is maximized \citep{kim_oup_2007}.

In the literature, both methods are often run on a relatively small range of ranks around the point at which the authors expect the optimal rank to be.
With cophenetic correlation, the optimal rank is chosen based on an extremely small drop in the coefficient over the range.
In fact, one run in \citet{brunet_nas_2004} was shown to be inconclusive based on the lack of a drop on the short range of ranks tested.
In our testing, we find the behavior of cophenetic correlation to be erratic on all of our datasets when testing higher ranks than was tested in the original implementation.
Similarly, we find the dispersion coefficient to be generally increasing as a function of rank after the initial drop.
In both cases, these metrics leave the user unsure of what to pick unless choosing to run on a limited range of ranks.
It is not always possible to determine an appropriately small range to check, especially when the rank of the underlying data cannot be determined \textit{a priori}.
Furthermore, the number of ranks included when plotting the cophenetic coefficient changes how stretched the plot is left-to-right between consecutive ranks, which drastically affects how steep a drop appears to be.

\subsubsection{Self-Comparison Methods}
Self-comparison methods can be subset into two categories, split validation and permutation comparison.
Split validation cuts the input matrix in half randomly, reorders the halves by the similarity of the basis vectors, then computes the similarity between the two halves.
This similarity metric can take a few forms but has shown success in the past using adjusted Rand index (ARI) \citep{hubert_joc_1985,grossberger_plos_2018} and inner product \citep{sotiras_pnas_2017}.
We choose not to compare against inner product as it had poor performance on real data in the testing performed by \citet{muzzarrelliI_IJCNN_2019}.

Permutation compares the slope of the elbow of the reconstruction error of the factorization of $\mA$ against the slope of the elbow of the reconstruction error of a permuted version of $\mA$.
Effectively, this compares the ability of NMF to reconstruct the dataset against the ability of NMF to reconstruct a random matrix of exactly the same magnitude as the original matrix.
Effectively, when the slope of the reconstruction error of $\mA$ is equal to that or greater than the slope of the permuted matrix, no extra information is able to be extracted from the original dataset compared to a random one.

\subsubsection{Non-Categorized Methods}
The elbow method is a popular technique for rank determination used in cluster analysis.
This method involves plotting the residual as a function of $k$ and picking the elbow of the curve as the correct number of clusters to use.
The elbow in the graph is where the rate of decrease changes, representing the point at which increasing the number of clusters does not significantly improve the fit of the model.
Although the elbow method is at least partially subjective, we compare against it for completeness.

Akaike information criterion (AIC) was successful in determining rank on time-series data in \citet{cheung_embc_2015} using a modified implementation of NMF.
We choose not to compare against AIC as it had poor performance in \citet{gilad_iop_2020} and was otherwise criticized by \citet{ito_siam_2016} for requiring assumptions that do not necessarily hold in NMF.
In this study, we focus on a general approach that does not make statistical assumptions regarding NMF or the underlying data.

Similarly, we do not compare against \citet{cai_semantic_2022}, a sequential hypothesis testing method, due to their requirement that the underlying data follows certain distributions.

For similar reasons, we do not compare against the category of Bayesian methods due to their requirements for \textit{a priori} knowledge of the distribution for prior estimation \citep{schmidt_icass_2009,cemgil_cin_2009}.

Additionally, we do not compare against methods utilizing minimum description length (MDL) as they assume a statistical model of the NMF \citet{yamauchi_springer_2012}.

Relevance determination is worthy of mention here but is ultimately not relevant to the discussion in this paper.
Essentially, these methods identify relevant clusters given a larger rank NMF decomposition \citet{tan_spars_2009}.

\subsubsection{Cross-Validation-Based Approaches}

A variety of methods for NMF rank determination using cross-validation (CV) have been proposed in the literature. These include bi-CV by \citet{owen_ims_2009}, and imputation-based CV by \citet{kanagal_semantic_2010}.
We choose not to compare against bi-CV as the results have been shown in practice to be unclear or unstable by \citet{kanagal_semantic_2010, gilad_iop_2020}.

Imputation-based CV is performed by optimizing for $\mW$ and $\mH$ given an imputed version of $\mA$ in which a percentage of values are denoted as missing \citep{kanagal_semantic_2010}.
The use of a Wold holdout pattern has been shown to be performant and is most widely used \citep{wold_technometrics_1978, kanagal_semantic_2010}.
We denote withheld values as $1$ in the binary masking matrix, $\mM$, and $0$ otherwise.
These values are hidden from computation during optimization. In most implementations, the reconstruction error is afterward calculated as
\begin{equation} 
    \frac{\norm{\mM \odot \left(\mA - \mW \mH\right)}_F^2}{\norm{\mM}_F^2}.\label{eq:ReconstructionError}
\end{equation} 
In the implementation originally proposed by \citet{kanagal_semantic_2010}, this is performed multiple times per rank and then averaged; the rank with lowest mean reconstruction error is chosen.
For the purpose of comparison in this study, we will call this method KS-CV, based on the last initials of the authors.

A variety of methods have been proposed that directly build on the work of \citet{kanagal_semantic_2010} such as MADImput, MSEImput, and CV2K \citep{muzzarrelliI_IJCNN_2019,gilad_iop_2020}.
Each of these three methods optimizes over every rank of interest a number of times, as was performed by \citet{kanagal_semantic_2010}.
For MADImput, the Median Absolute Deviation (MAD) of the reconstruction errors is calculated at each rank and the rank with lowest MAD is chosen \citep{muzzarrelliI_IJCNN_2019}.
We choose not to compare against MSEImput because it performed poorly on all but simulated data \citep{muzzarrelliI_IJCNN_2019}.
Differing from the other imputation methods described, the authors of \citet{gilad_iop_2020} compute the error as the $\normlone$-norm of the error over the masked values, computed against a normalized version of the initial matrix which allows them to normalize both $\mW$ and $\mH$, as detailed in Algorithm 2 in their paper.
The rank with minimum median reconstruction error calculated as stated is chosen, but is adjusted down based on a correction step determined by a Wilcoxen rank-sum test \citet{gilad_iop_2020}.

\subsection{Complexity}

The majority of rank determination methods compute NMF using a standard optimization algorithm a number of times, then compute their metric.
While the cost of computing these metrics is not free, it is typically substantially less than the cost of computing the NMF decomposition.
On the other hand, imputation-based methods must deal with missing values during the computation of the NMF decomposition, fundamentally altering how the NMF decomposition is computed.  The complexity of imputation-based CV methods is often overlooked but is a significant burden in practice.

Before discussing the complexity of computing an NMF decomposition with missing values, we must first discuss the complexity of a standard NMF decomposition.
As is common practice, we only count multiplication and division, and we assume naive algorithms for common operations such as matrix multiplication.
The amount of work performed at a given rank for a single optimization iteration using SCD when no values are missing is described by
\begin{equation}
    2mnk + 2mk^2 + 2nk^2.\label{eq:originalComplexity}
\end{equation}
This assumes two Gram matrix computations, one computation of $\mB_W$ and $\mB_H$, and one computation of $\mW\mG_H$ and $\mH^T\mG_W$.
Under the assumption that $k \ll \min(m, n)$, the time complexity is $\mathcal{O}(mnk)$.
The amount of work required for a single iteration of NMF with MU is slightly more, but results in the same overall time complexity.

The computation of CV with missing values is considerably more complex.
The implementation provided by \citet{lin_springer_2020}, and used by \citet{gilad_iop_2020}, implements imputation-based CV by creating a new Gram matrix for each row or column affected by missing values.
That is to say that each column, $\mH_{:,j}$, in the update of $\mH$ requires a different Gram matrix, denoted by $\mG_W^{(j)}$, and created as
\begin{align*}
    \mG_W^{(j)} &= \left(\text{diag}(\oldneg\mM_{:, j}) \mW\right)^T \left(\text{diag}(\oldneg\mM_{:, j}) \mW\right)\nonumber\\
                &= \mW^T\text{diag}(\oldneg\mM_{:, j})\mW.\\
\intertext{Similarly, each row, $\mW_{i,:}$, in the update of $\mW$ requires a different Gram matrix, $\mG_H^{(i)}$, which is created as}
    \mG_H^{(i)} &= \left(\mH \text{diag}(\oldneg\mM_{i, :})\right)\left(\mH \text{diag}(\oldneg\mM_{i, :})\right)^T\nonumber\\
    &= \mH \text{diag}(\oldneg\mM_{i, :}) \mH^T,
\end{align*}
where $\oldneg$ denotes element-wise logical negation.
Taking into account the need to compute all of these Gram matrices, the missing value computation cost per iteration is captured by
\begin{equation}
    2mnk^2 + 2mnk + mk^2 + nk^2.\label{eq:CVComplexity}
\end{equation}
This assumes the computation of $m + n$ Gram matrices, one computation of $\mB_W$ and $\mB_H$, one computation of $\mW_{i,:}\mG_H^{(i)}$ for all $i \in \{ 1, 2, \dots, m \}$ and $(\mH_{:,j})^T\mG_W^{(j)}$ for all $j \in \{ 1, 2, \dots, n \}$, and at least one missing value per row and column.
This gives a time complexity of $\mathcal{O}(mnk^2)$, but attention should be given to the two equations which model their behavior.
The dominating term of \autoref{eq:originalComplexity} is contained within \autoref{eq:CVComplexity}, meaning that nearly all of the work required to compute the original factorization must be performed in addition to the additional gram matrix related work.
For datasets of relatively small size, this is not particularly burdensome, but it is entirely unfeasible on larger datasets as discussed later in \autoref{sec:ResultsText}.
It should be noted that the CV2K implementation from \citet{gilad_iop_2020} has lower time complexity but necessitates the storage of five additional matrices of equal size to $\mA$; this is similarly burdensome for matrices of sufficient size.


\section{Residual Sensitivity to Initial Conditions (RSIC)}\label{sec:Method}

As previously described in \autoref{sec:Limitation}, NMF is highly sensitive to the initial conditions of the $\mW$ and $\mH$ matrices.

\begin{figure}[ht]
    \centering
    \includegraphics[width=\textwidth]{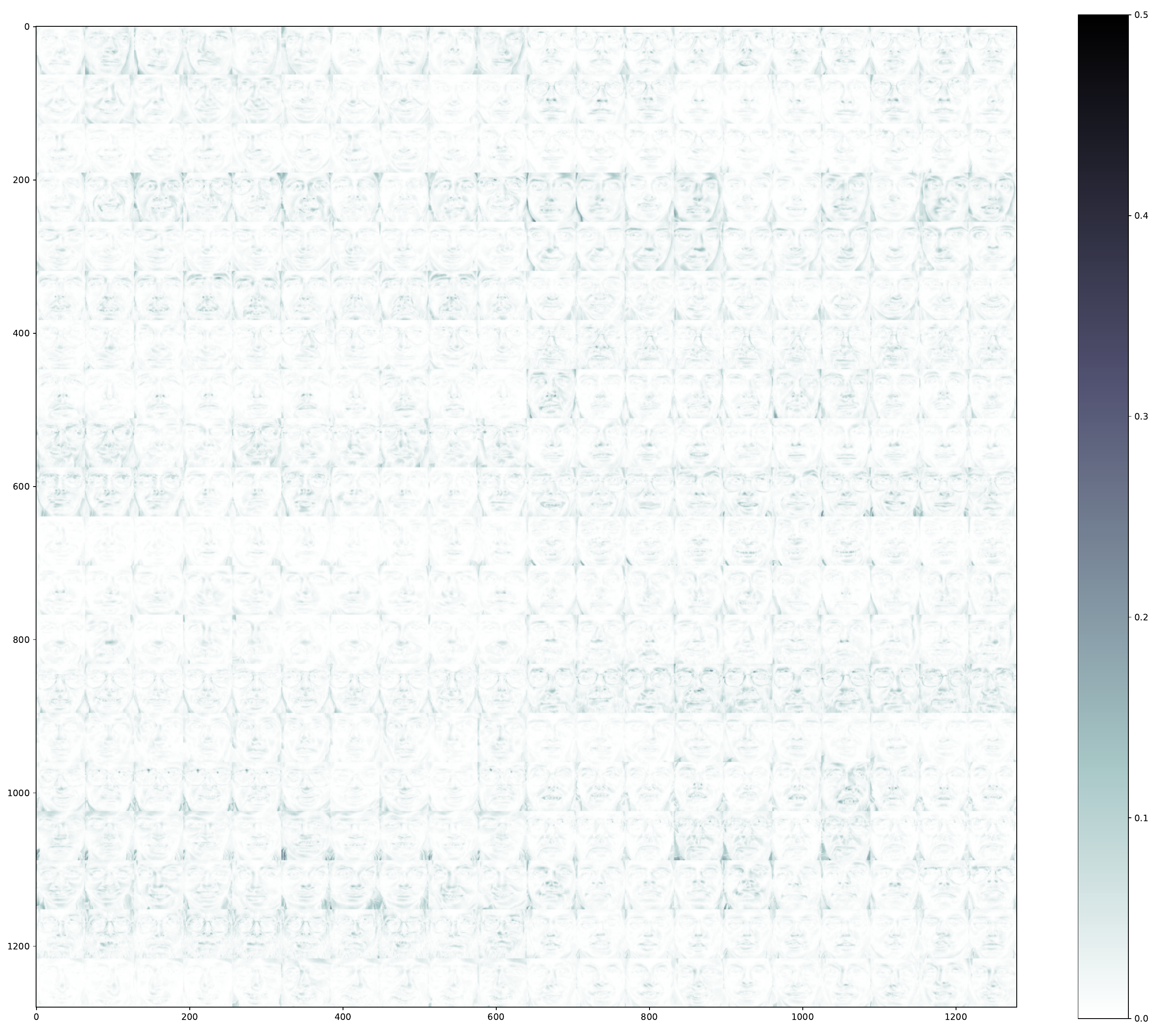}
    \caption{The delta between the smallest and largest error at each point in the rank-10 reconstruction of the first face in the Faces dataset.}
    \label{fig:olivetti_variances}
\end{figure}

We have found that, even amongst factorizations whose residuals have comparable Frobenius norms, the reconstruction error computed at any point may be arbitrarily worse or better in any given factorization.
This observation highlights the inherent unpredictability in NMF due to random initializations and can be seen clearly in \autoref{fig:olivetti_variances}.
This figure shows the delta between the smallest and largest error at each point in the rank-10 reconstruction of the first face in the Faces dataset (described in detail later in \autoref{subsub:Faces}).
The maximum delta figure is plotted over the models which had within $10\%$ Frobenius norm of the residual of the median residual model.
This figure shows that, even within models that have comparable norms, there exists massive deviation in the reconstruction error when considered coordinatewise.
Similar behavior was found to be present in all of the datasets that were tested.

Of interest, these observations reveal a pattern within the variability: at certain ranks, the sensitivity to initial conditions as measured by the deltas in reconstruction error diminishes greatly, forming what can be described as ``islands of stability''.
These ranks, where the variance in reconstruction error is minimal despite different initializations, stand out against nearby ranks that exhibit high sensitivity.

We hypothesize that these ``islands of stability'' are not merely random occurrences but could be indicative of inherent structure or patterns within the data that are particularly well-captured by NMF at these specific ranks.
These stable ranks may correspond with factorizations that more effectively distill the essential features of the data and are less influenced by the initial conditions.

To investigate this, we measure this by computing a number of factorizations at each rank, where the randomly generated initial conditions are related across ranks in a scheme described next in \autoref{sub:randominit}.
We then attempt to quantify the Residual Sensitivity to Initial Conditions (RSIC) at a given rank using the metric developed in \autoref{sub:mci}.
Plotted as a function of the rank, $k$, we then identify ``islands of stability" as potential ranks of interest that should be further investigated.

\subsection{Progressive Random Initialization}\label{sub:randominit}
We implement a progressive random initialization scheme, which allows individual random initializations to be related across ranks, smoothing out variations in reconstruction error across ranks of the NMF factorization for a single progressive random initialization.
For a given initialization with maximum rank of interest, $k_{\text{max}}$, let $\rmW_{\text{init}}$ and $\rmH_{\text{init}}$ be random matrices of dimension $m \times k_{\text{max}}$ and $k_{\text{max}} \times n$, respectively, whose entries are generated from the uniform distribution on the half interval $[0,1.0)$.
Assume $a$ is the desired number of initializations per rank.
Let $\rmW_{\text{init}}^{(r)}$ be the $r$-th random initialization of $\rmW_{\text{init}}$, and $\rmW_\text{init}^{(r, k)}$ is a copy of the left submatrix $\left(\rmW_\text{init}^{(r)}\right)_{:,:k}$.  

Likewise for $\rmH_\text{init}$, let $\rmH_\text{init}^{(r, k)}$ be a copy of the upper submatrix $\left(\rmH_\text{init}^{(r)}\right)_{:k,:}$.
Then, for a given initialization, $r$, and letting $k_\text{min}$ be the minimum rank of interest, we have the ordered set of tuples of initial matrices,
\begin{align*}
        \sS_{\text{init}}^{(r)} &= \left\{ \left(\rmW_{\text{init}}^{(r,k_\text{min})}, \rmH_{\text{init}}^{(r,k_\text{min})}\right), \left(\rmW_{\text{init}}^{(r,k_\text{min} + 1)}, \rmH_{\text{init}}^{(r,k_\text{min} + 1)}\right), \dots, \left(\rmW_{\text{init}}^{(r,k_\text{max})}, \rmH_{\text{init}}^{(r,k_\text{max})}\right) \right\}.
\end{align*}
Assume a set, $\sS^{(*)}_\text{init}$, exists for each progressive random initialization we are performing.
For simplicity, we index these sets as
\begin{align*}
    \sS^{(r,k)}_\text{init} &= \left(\rmW_{\text{init}}^{(r,k)}, \rmH_{\text{init}}^{(r,k}\right).
\end{align*}
Given this indexing, this specifically means that the matrices in $\sS^{(*, k-1)}_\text{init}$ contain a copy of the respective matrices in $\sS^{(*, k)}_\text{init}$ for all $k \in \{k_\text{min} + 1, k_\text{min} + 2, \dots, k_\text{max}\}$.
This is such that $\rmW_\text{init}^{(*,k-1)}$ contains a copy of the left submatrices of all but the last column of $\rmW_\text{init}^{(*,k)}$ and $\rmH_\text{init}^{(*,k-1)}$ contains a copy of the upper submatrices of all but the last row of $\rmH_\text{init}^{(*,k)}$.

Finally, let
\begin{equation*}
    \sS_{opt}^{(r,k)} \longleftarrow \text{NMF}\left(\mA, \sS_{init}^{(r,k)} \right)
\end{equation*}
be the resulting factorization of $\mA$ after performing NMF with a given rank and initialization pairing.
This factorization transforms the unoptimized matrices, giving their optimized counterparts.
This is performed over all progressive random initializations and ranks, giving $\sS_\text{opt}$.

\subsection{Mean Coordinatewise Interquartile Range (MCI)}\label{sub:mci}
The goal of this metric is to capture the sensitivity of the relative reconstruction error at a given rank by measuring the average spread of relative reconstruction errors at each point.
This is computed by taking the interquartile range (IQR) of the relative reconstruction error at each point over the number of initializations.

Let $\mR^{(k)}$ be a matrix of dimension $a \times mn$, and assume $\vectorize{*}$ flattens an $m \times n$ matrix to a $mn$ dimension vector in any consistent ordering.
The values of $\mR^{(k)}$ are given by
\begin{align}
    \mR_{r, :}^{(k)} &= \vectorize{\mA - \mW_\text{opt}^{(r,k)}\mH_\text{opt}^{(r,k)}
    },\nonumber
\end{align}
for all $r \in \{ 1, 2, \dots, a \}$ with given rank, $k$.
The Mean Coordinatewise IQR (MCI) at each rank, $k$, may now be computed as
\begin{align}
    \text{MCI}^{(k)} &= \frac{1}{mn} \sum_{j=1}^{mn}{\text{IQR}\left(\mR_{:, j}^{(k)} \right)},\nonumber
\end{align}
where IQR computes the interquartile range of a vector.
This is performed for all $k \in \{k_\text{min}, k_\text{min} + 1, \dots, k_\text{max}\}$.

Computing the MCI over a large number of factorizations requires the storage of all of these factorizations.
Although this may be a significant amount of space for large matrices, the total amount of work required is significantly less than those methods based on cross-validation techniques.
Additionally, the post-processing step is able to be computed in batches in a trivial manner, allowing for computation on machines with lower amounts of memory.

\section{Datasets}\label{sec:Datasets}
We define the ``true rank'' of a dataset as the number of classes in the underlying dataset.
Additionally, certain datasets cannot be designated a single true rank as sub-classes may exist which equally make sense to target in the analysis of a dataset.
We discuss any potential sub-classes in our discussion of each dataset.
This does not necessarily correlate directly to the optimal rank of an NMF decomposition.
We compare our method on eight datasets from three different disciplines.
We break this section into three subsections, one for each type of dataset used.
All datasets are formatted such that each row represents a sample and the columns within a row are features.
This means that the two single cell datasets, introduced next, are transposed from how they are generally presented in the literature.

\subsection{Single Cell Datasets}\label{sec:DatasetsSingle}
ALL-AML was originally described in \citet{golub_science_1999} and retrieved using the package provided by \citet{dataset_golub_package}.
This dataset is $38 \times 5000$ in dimension, consisting of the $5000$ most highly varying human genes in the original dataset and taken from $38$ bone marrow samples \citet{golub_science_1999,dataset_golub_package}.
Of these $38$ samples, $27$ are related to acute lymphoblastic leukemia (ALL) and $11$ acute myeloid leukemia (AML) \cite{golub_science_1999}.
Given the existence of two cancer types, the true rank is $2$, though most authors find a rank $3$ NMF decomposition to be more informative \citep{brunet_nas_2004}. 

The PBMC3K dataset was retrieved from \citet{dataset_genomic_2016} and has dimension $2700 \times 13714$.
This dataset consists of $2700$ cells and $13714$ genes and comes pre-filtered for relevance.
This dataset has no consensus true rank but has been shown using domain knowledge to contain 9 distinct cell types \citep{du_g3_2020}, which will be treated as the true rank for our purposes.

\subsection{Image-Based Datasets}\label{sec:DatasetsImage}

\subsubsection{Swimmer}
The Swimmer dataset was first described in \citet{donoho_neurips_2003} and retrieved from the package released by \cite{tan_spars_2009}.\footnote{The package may be found at \url{www.irit.fr/~Cedric.Fevotte/extras/pami13/ardnmf.zip}}
Swimmer has dimension $256 \times 1024$ and is a synthetic dataset consisting of $256$ flattened $32 \times 32$ black and white stick figure images meant to mimic a variety of breaststroke positions. There are a total of $16$ limb positions, giving a true rank of $16$.

\subsubsection{Full Digits \& Dig0246}
We retrieved the ``Optical Recognition of Handwritten Digits'' dataset by \citet{dataset_digits_1998} from the package by \citet{pedregosa_jmlr_2011}. We use this both as the full dataset and additionally use a selected subset for experimentation.
We call the full dataset Full Digits and the subset Dig0246, containing only the digits $\{0, 2, 4, 6\}$. 
We subset in this manner for the sake of comparison and consistency with the authors of \citet{muzzarrelliI_IJCNN_2019}, who did not run on the full dataset.
We should note that the clustering behavior of NMF appears to more clearly separate the classes in Dig0246 than with Full Digits.
An example of this clustering behavior will be shown later in \autoref{sec:ResultsImage}.
The true ranks are $10$ and $4$ for Full Digits and Dig0246, respectively.

\subsubsection{Faces}\label{subsub:Faces}
We retrieved the ``AT\&T Laboratories Cambridge Faces'' dataset by \cite{dataset_database_of_faces_1994} using the package provided by \citet{pedregosa_jmlr_2011}. This dataset has dimension $400 \times 4096$, consisting of $400$ flattened black and white images of size $64 \times 64$.
The dataset consists of $10$ people, with $40$ images from each.
Given the number of subjects, the true rank is $10$.
The images were originally $92 \times 112$ in dimension but have been modified in the version retrieved from \citet{pedregosa_jmlr_2011}.

\subsection{Text-Based Datasets}\label{sec:DatasetsText}

All text datasets were transformed using a bag-of-words model, \texttt{CountVectorizer}, provided by \citet{pedregosa_jmlr_2011}.
The maximum document frequency was set to $95\%$, minimum document frequency to an integer value of $2$, and the stop-words set to \texttt{english} as provided by the implementation.
The maximum features was set to $4000$ and $30000$ for NewsGroup4000 and Web of Science, respectively.
All other options were left to their defaults.

\subsubsection{NewsGroup4000}
We obtained the NewsGroup4000 dataset, originally found at \citep{dataset_newsgroup_1997}, using the packaged provided by \citet{pedregosa_jmlr_2011}.
This dataset is of dimension $11314 \times 4000$ and contains $20$ topics ranging from sports to medical. Given the number of topics, the true rank is $20$.

\subsubsection{Web of Science}
We obtained the Web of Science dataset, which was originally described in \citet{kowsari_icmla_2017} from \citet{dataset_web_of_science_2017}.
The data was preprocessed as previously described using \texttt{CountVectorizer}.
This dataset is of dimension $11967 \times 28095$ and contains $11967$ documents from $35$ categories and $7$ parent categories.
Fundamentally, there are two different true ranks in this dataset, $35$ and $7$.

\section{Experimental Setup}\label{sec:ExperimentalSetup}
We opt to use only publicly available packages for all significant computations.
Due to the large variety of datasets and methods we compare, for computational feasibility, we perform $100$ random initializations for each method on each dataset.
For each dataset, we let $k_\text{min} = 2$ and $k_\text{max} =  \min(m, n, 64)$ and perform each method on each rank between $k_\text{min}$ and $k_\text{max}$ inclusively.
Since it has been shown that MU converges to a solution more slowly than SCD \citep{lin_springer_2020}, SCD was used as the optimization routine for all methods in which the option was available.
We force all methods using SCD to run to $100$ optimization iterations per initialization and rank by setting the tolerance to $1\mathrm{e}{-16}$.
For the remaining methods in which SCD was not an option, MU was selected and forced to run for $500$ optimization iterations due to its slower convergence. 
For consistency with \cite{muzzarrelliI_IJCNN_2019}, all methods optimized the Frobenius norm defined in \autoref{eq:Euclidean}.
Before the start of computation for each method and dataset, the random state was set to the seed $123456789$.
For the computation performed with the packages provided by \citet{gilad_iop_2020, lin_springer_2020}, which offer multi-threading support, computation was performed on a dedicated workstation with a $32$ core Threadripper processor and $64$ GB of memory.
All other computation was performed sequentially on a consumer-grade Intel processor.

\subsection{Elbow}
For the comparison against the elbow method, we determine the elbow based on the average reconstruction error of each rank.
This average is computed based on the progressive random initialization scheme described earlier.
The results are provided based on a visual determination of where the bend appears to be.

\subsection{Cophenetic Correlation \& Dispersion Coefficient}
For the comparison with cophenetic correlation and dispersion coefficient methods, \citep{brunet_nas_2004,kim_oup_2007}, the NIMFA package provided by \citet{zitnik_jmlr_2012} was used.
This package does not have support for SCD and MU was used instead.
The initialization type was set to \texttt{random}, number of initializations to $100$, number of iterations to $500$, the update to \texttt{euclidean}, and objective function to \texttt{fro}.
We passed in the range of ranks from $k_\text{min}$ to $k_\text{max}$, and plotted the results, which are the cophenetic correlation coefficient and dispersion coefficient as determined by the package.
The optimal rank for each is then chosen based on the criteria set forth by the original authors in \cite{brunet_nas_2004} and \cite{kim_oup_2007} and briefly described in this paper in \autoref{subsub:Consensus}.

\subsection{Permutation}
For the comparison with permutation, we performed the optimization of NMF using the implementation provided by \cite{pedregosa_jmlr_2011}.
We shuffled the the columns individually for each row using the package provided by \citet{harris_nature_2020}, then optimized the permuted matrix separately.
The Frobenius norm of both the permuted and non-permuted matrix was computed as a function of rank.
Using SciPy's implementation of the Savitzky-Golay filter \cite{virtanen_nature_2020}, we approximate the slope of the residuals as a function of rank.
We look for the point where the reconstruction error of the unpermuted matrix decreases less sharply than that of the permuted matrix.
This corresponds to the point at which the slope of the residuals of the non-permuted matrix is greater than or equal that of the slope of the residuals of the permuted matrix because both are decreasing and the slopes are negative.
In terms of the approximated derivative, the rank selected is the point immediately before overtaking the elbow of the permuted matrix, or the point at which they are exactly equal. Due to floating point arithmetic, equality is considered a relative tolerance of $1 \times 10^{-8}$, computed relative to the larger of two derivative estimates.
This is performed $100$ times and the median result is computed.

\subsection{ARI Method}
For the comparison against the ARI method, we randomly split the matrix into two equal parts along the dimension relating to factors.
If splitting into even parts is not possible, the larger split is truncated to size.
Then NMF is computed on each split using the implementation provided by \citet{pedregosa_jmlr_2011}.
The cosine distance between factors is computed using the package provided by \citet{virtanen_nature_2020}, giving a distance matrix.
This is then passed to the \texttt{linear\_sum\_assignment} function as provided by \citet{virtanen_nature_2020}, which implements the Hungarian algorithm described in \citet{crouse_taes_2016}.
The result is the ARI, which is then averaged over all initializations at a given rank.
The rank with highest mean ARI is selected.
We note that we ran with both the mean and median ARI, which resulted in the same decided rank in all cases.

\subsection{Cross Validation Methods}
For the comparison against KS-CV and MADImput, we use the package provided by \citet{lin_springer_2020}.
The output of this package is the reconstruction error at each rank as previously defined in \autoref{eq:ReconstructionError}.
Using this output, we are able to compute each of these metrics and choose a rank based on where it is minimized.
For the comparison against CV2K, we use the package provided by the author \citet{gilad_iop_2020}, which returns both a similar output and the chosen rank based on their criteria.

We note that we have chosen to use a different initialization scheme than was used by the authors of \citet{gilad_iop_2020,muzzarrelliI_IJCNN_2019}.
In \citet{muzzarrelliI_IJCNN_2019}, the authors performed $100$ runs, initializing $20$ times in each run before choosing the best model as defined by reconstruction error.
In \citet{gilad_iop_2020}, the authors initialize a large number of times, optimizing for hundreds of iterations, and chose the best model defined by reconstruction error to continue.
In order to perform this, we would need to run $2000$ times for the \citep{muzzarrelliI_IJCNN_2019} method and $10000$ times for the \citep{gilad_iop_2020} method.\footnote{A number of initialization was not described in the CV2K paper but was instead found on the Github linked in the paper.}
In addition to the considerable length of time required, we do not believe it is fair to arbitrarily run one method an order of magnitude more times than another.
Instead, we focus on running all methods for the same number of initializations.
For methods that rely on differences across random initializations, this means that they do not get the added benefit of testing against thousands of initializations.

In order to enforce these requirements, a tolerance may be provided to the \citep{lin_springer_2020} package.
For the \citep{gilad_iop_2020} package, minor modifications to the code are required.
The code is modified to force running all 100 iterations and the \texttt{init\_factor\_matrices} function is modified to return the first generated $\mW$ and $\mH$.

\section{Results}\label{sec:Results}

In this section, we present results and compare against other methods.
In our analysis of the PBMC3k, NewsGroup4000, and Web of Science datasets, significant computational hurdles were encountered.
Due to RAM constraints on our system, we were forced to run CV2k with only $16$ threads for both the PBMC3k and NewsGroup4000 datasets.
This limitation resulted in an excessively prolonged runtime of over 30 days for PBMC3k and longer for NewsGroup4000.
Given the size of the Web of Science dataset, a further reduction in threads would have been necessary, leading to an even longer computational burden.

NNLM-CV was not constrained by memory requirements but was unable to complete even a single run on the PBMC3k dataset in $24$ hours, meaning the full $100$ runs would take over $100$ days.
The larger scales of NewsGroup4000 and Web of Science implied that completion times would be significantly longer.
This difficulty, owing to the incredible time complexity of the method as described in \autoref{eq:CVComplexity}, led to the conclusion that the burden of computation for these methods is infeasible.

Consequently, the result table for these methods on the mentioned datasets are denoted as ``N/A''.
In other instances, where the output was inconclusive, we have designated the results ``undetermined'' (Und.).
For instance, undetermined would be noted when using the elbow method where there is no clear elbow or when using the permutation method but the slope of the permuted is steeper than the slope of the unpermuted data from the beginning.
The remaining results are presented either as integer values or in an increasing order set format, which is applicable only to permutation as well as MCI-RSIC as previously described.

\subsection{Single Cell Datasets}\label{sec:ResultsSingle}

\begin{figure}[ht]
    \centering
    \includegraphics[width=\textwidth]{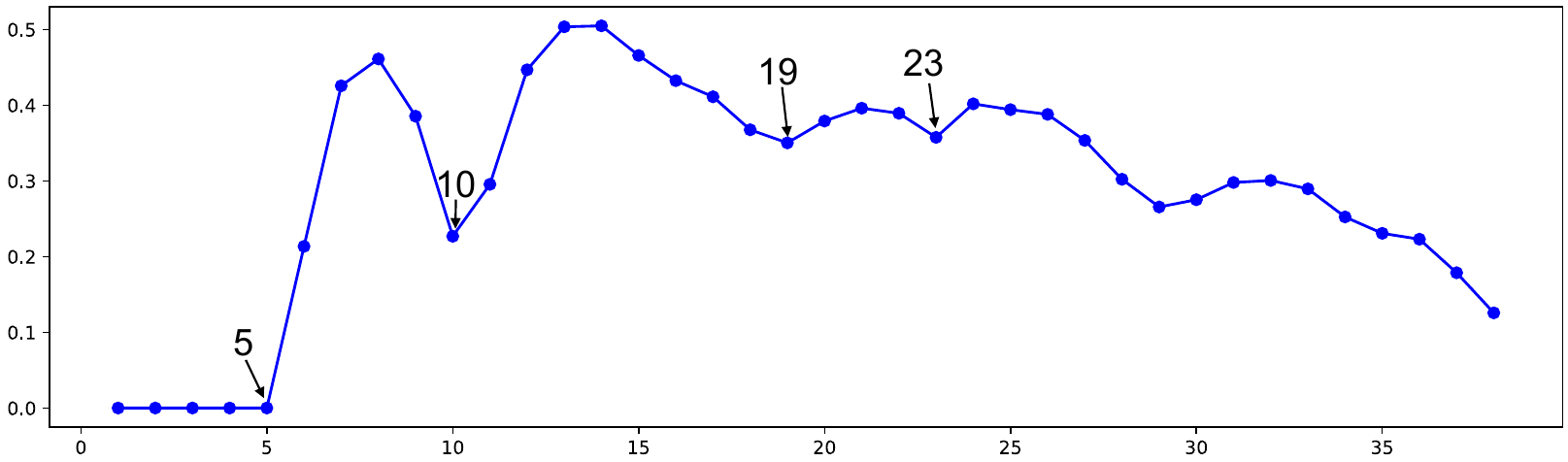}
    \caption{Mean Coordinatewise IQR (MCI) ($y$-axis) vs rank ($x$-axis) for the ALL-AML dataset for ranks 1 through 38.  We identify ranks 5, 10, 19, and 23 as ``islands of stability'' and thereby potential ranks of interest.}
    \label{fig:golub-iqr}
\end{figure}

We show the RSIC-MCI metric as a function of rank on the ALL-AML dataset in \autoref{fig:golub-iqr}.
The horizontal axis, representing rank, ranges from $1$ to $38$.
Based on our selection criteria described before, we find rank $5$ to be the first island of stablility.
Note that ranks $1$ through $5$ all have the same or similar MCI values, but rank $5$ is the first rank before a significant increase in MCI.

We found that our method and permutation performed poorly on the ALL-AML dataset, whereas cophenetic, dispersion, and CV2k all converged on the generally agreed upon rank and ARI and KS-CV converged on the true rank.
For PBMC3k, no method returned a solution that contended with the true rank although MCI-RSIC and elbow both returned nearby ranks.

\begin{table}[h!]
\centering
\caption{Results for all evaluated rank determination methods in comparison to true rank on the ALL-AML and PBMC3K single cell datasets.}
\begin{tabular}{|c|c|c|}
\hline
Method & ALL-AML & PBMC3k \\
\hline
MCI-RSIC & $\{5,~10,~19,~23\}$ & $\{3,~7,~11\}$\\
Elbow & $4$ & $8$ \\
Cophenetic & $3$ & $2$ \\
Dispersion & $3$ & $2$ \\
Permutation & $5$ & $41$ \\
ARI & $2$ & $2$ \\
KS-CV & $2$ & N/A \\
CV2K & $3$ & N/A \\
MADImput & $4$ & N/A \\
\hline \hline
True Rank & $2$ & $9$\footnote{Recall that this dataset has no consensus true rank.} \\
\hline
\end{tabular}
\end{table}

\subsection{Image-Based Datasets}\label{sec:ResultsImage}

\begin{figure}[ht]
    \centering
    \includegraphics[width=\textwidth]{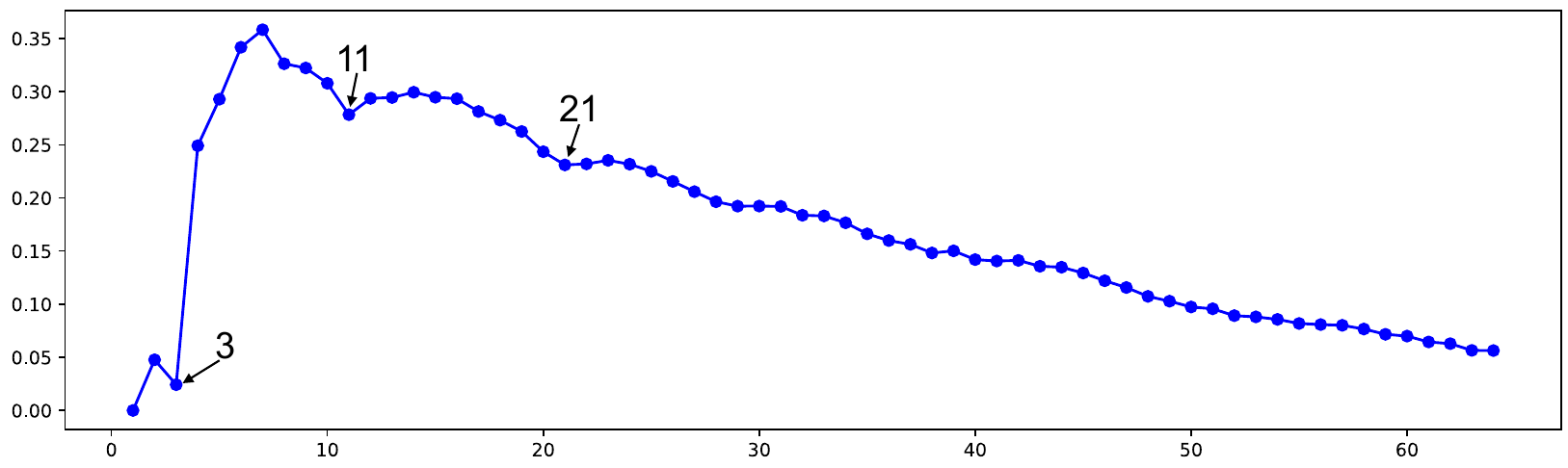}
    \caption{Mean Coordinatewise IQR (MCI) ($y$-axis) vs rank ($x$-axis) for the Full Digits dataset for ranks 1 through 64.  We identify ranks 3, 11, and 21 as ``islands of stability'' and thereby potential ranks of interest.}
    \label{fig:fulldigit-iqr}
\end{figure}

We show the RSIC-MCI metric as a function of rank on the Full Digits dataset in \autoref{fig:fulldigit-iqr}.
This image plots the MCI metric from rank $1$ to $64$.
Like the other methods tested, this method performed equivalently poorly on the Full Digits dataset when considering the true rank.
There is significant mixing between the ranks, which is indicative of poor clustering behavior.

\begin{figure}[ht]
    \centering
    \includegraphics[width=\textwidth]{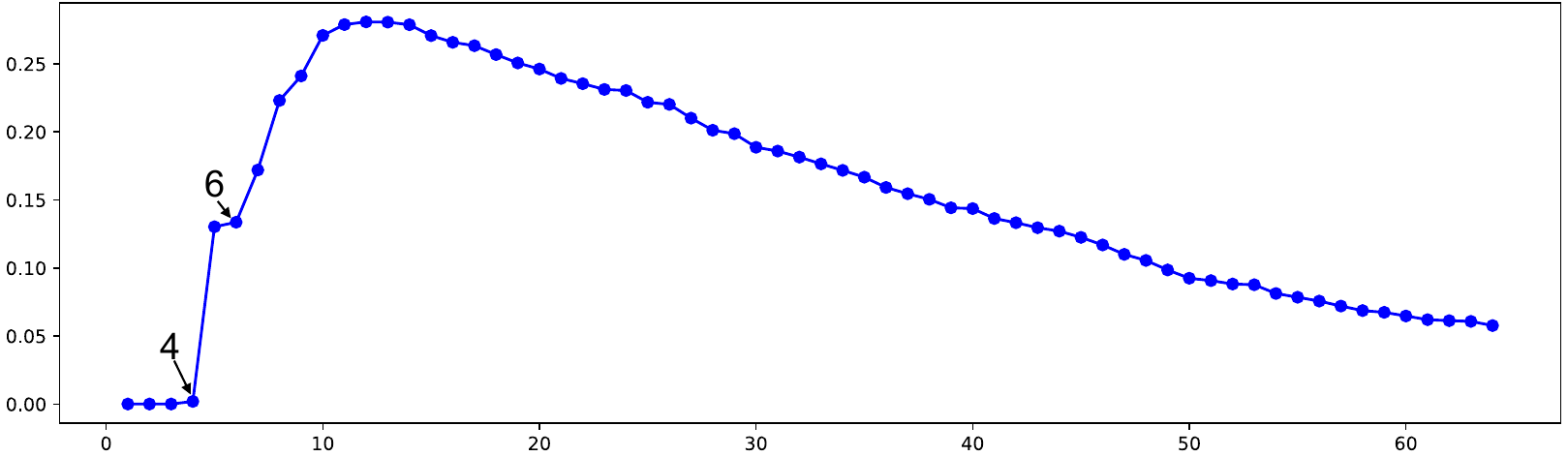}
    \caption{Mean Coordinatewise IQR (MCI) ($y$-axis) vs rank ($x$-axis) for the Dig0246 dataset for ranks 1 through 64.  We identify ranks 4 and 6 as ``islands of stability'' and thereby potential ranks of interest.}
    \label{fig:lessdigits-AvgIQRRelative.pdf}
\end{figure}

In addition to the RSIC-MCI metric on Full Digits, we show the metric on Dig0246 in \autoref{fig:lessdigits-AvgIQRRelative.pdf}.
This image plots the RSIC-MCI metric from rank $1$ to $64$.
This clearly shows an island of stability at rank $4$ and provides evidence for on at rank $6$.

\begin{figure}[ht]
    \centering
    \includegraphics[width=\textwidth]{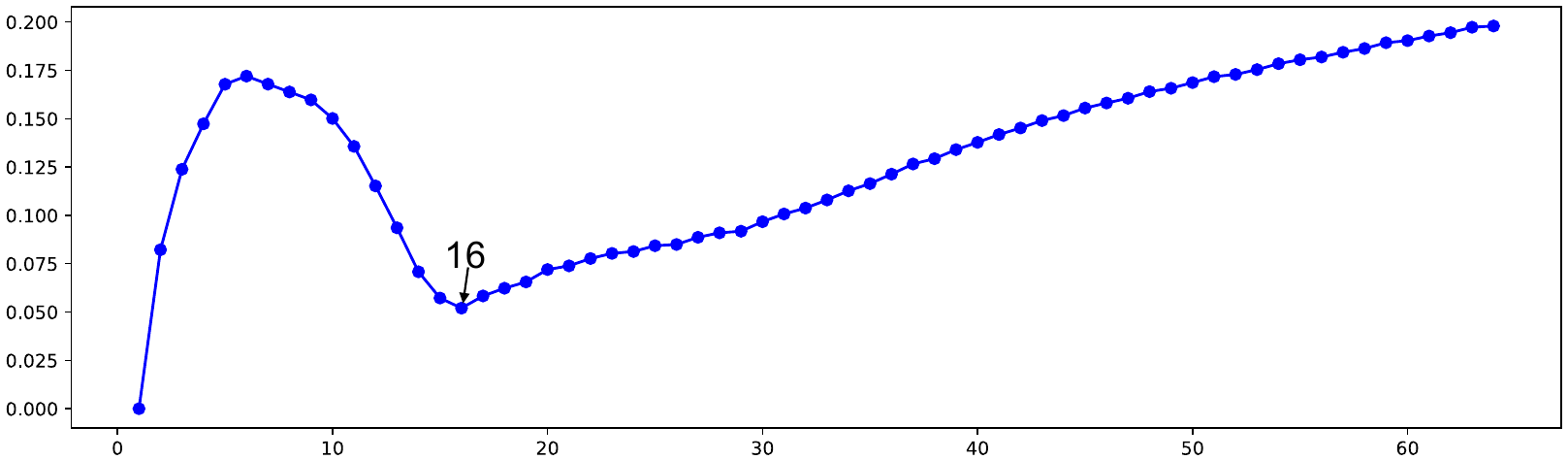}
    \caption{Mean Coordinatewise IQR (MCI) ($y$-axis) vs rank ($x$-axis) for the Swimmer dataset for ranks 1 through 64.  We identify rank 16 as the only ``island of stability'' and thereby potential rank of interest.}
    \label{fig:swimmer-iqr}
\end{figure}

Finally, we show the RSIC-MCI metric as a function of rank on the Swimmer dataset in \autoref{fig:swimmer-iqr}.
This image plots the MCI metric from rank $1$ to $64$ and clearly shows an island of stability at rank $16$, which corresponds to the true rank of the dataset.

\begin{figure}[ht]
    \centering
    \begin{subfigure}[b]{\textwidth}
        \includegraphics[width=\textwidth]{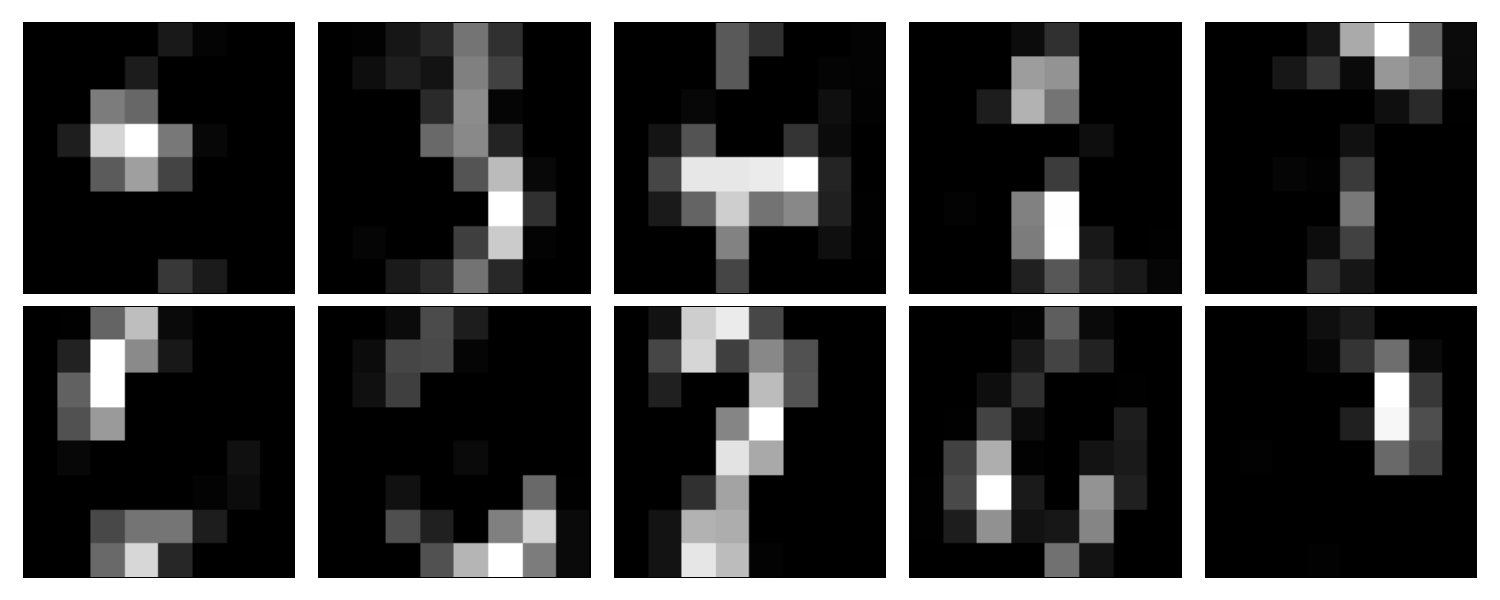}
        \caption{Rank 10}
        \label{fig:fulldigit-rank10}
    \end{subfigure}\\
    \begin{subfigure}[b]{\textwidth}
        \includegraphics[width=\textwidth]{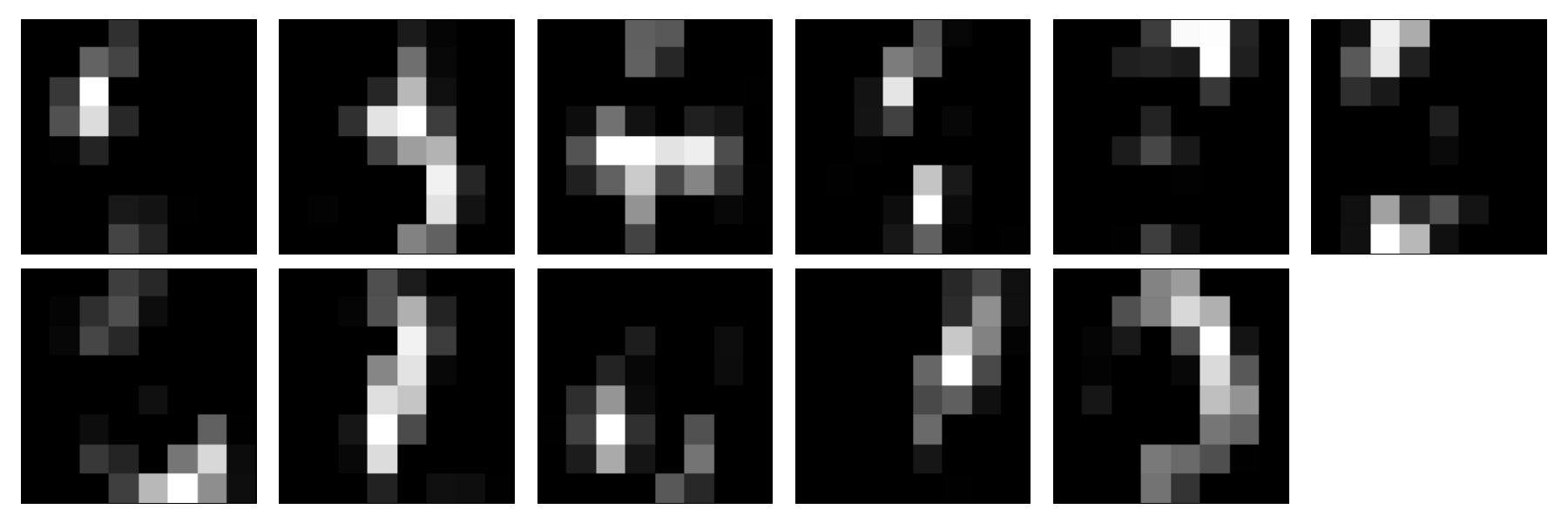}
        \caption{Rank 11}
        \label{fig:fulldigit-rank11}
    \end{subfigure}
    \caption{Clustering behavior of the Full Digits dataset at ranks 10 and 11.}
    \label{fig:fulldigit-clustering}
\end{figure}

\autoref{fig:fulldigit-clustering} shows poor clustering behavior at the true rank of $10$ on the Full Digits dataset.
Additionally, we show the clustering behavior at rank $11$, which is a nearby rank that we suggest.
This figure shows that the clustering behavior at both ranks is rather poor, and provides evidence for why all methods performed poorly on this dataset.
Indeed, there appears to be no clear separation between the classes in the dataset.
This poor clustering behavior was present in each rank we tested, and is indicative of the poor performance of all methods on this dataset.

For the image based datasets, MCI-RSIC and elbow returned the true rank on the Swimmer dataset. The permutation method consistently detected a rank greater than the true rank, across all image datasets.
For the faces dataset, MCI-RSIC performed poorly in terms of the true rank and only elbow returned the true rank.
On Full Digits, all methods performed poorly though the clustering behavior of this dataset is poor as seen in \autoref{fig:fulldigit-clustering}.
For Dig0246, MCI-RSIC, elbow, cophenetic, dispersion, and ARI all returned the true rank.
We note that MADImput found the true rank in their paper \cite{muzzarrelliI_IJCNN_2019} and the discrepancy here is most likely due to their use of 1900 more initializations than allowed in our study.

\begin{table}[h!]
\centering
\caption{Results for all evaluated rank determination methods in comparison to true rank on Swimmer, Faces, Full Digits, and Dig0246 image datasets.}
\begin{tabular}{|c|c|c|c|c|}
\hline
Method & Swimmer & Faces & Full Digits & Dig0246 \\
\hline
MCI-RSIC & $16$ & $\{2,~6,~9,~18\}$ & $\{3,~11,~21\}$ & $\{4,~6\}$ \\
Elbow & $16$ & $10$ & $5~(\text{Und.})$ & $4$ \\
Cophenetic & Und. & $2$ & $2$ & $4$ \\
Dispersion & $64+$ & $2$ & $2$ & $4$ \\
Permutation & $18$ & $50$ & $22$ & $17$ \\
ARI & $13$ & $2$ & $5$ & $4$ \\
KS-CV & $14$ & $51$ & $12$ & $12$ \\
CV2K & $17$ & $64$ & $16$ & $16$ \\
MADImput & $13$ & $60$ & $2$ & $11$ \\
\hline \hline
True Rank & $16$ & $10$ & $10$ & $4$ \\
\hline
\end{tabular}
\end{table}

\subsection{Text-Based Datasets}\label{sec:ResultsText}

No method was able to return the true rank for NewsGroup4000 although MCI-RSIC and dispersion both return a nearby rank. The others underestimate more substantially (elbow, cophenetic, dispersion, ARI), are infeasible to run (KS-CV, CV2K, MADImput), or failed to identify any rank (permutation).

For Web of Science, MCI-RSIC returns the true rank for the category as well as a nearby rank to the number of subcategories.  Elbow detects a nearby rank for the category.  All other methods either substantially undershoot (cophenetic, dipsersion, ARI), are infeasible to run (KS-CV, CV2K, MADImput), or indicated the rank was greater than the tested range (permutation).

\begin{table}[h!]
\centering
\caption{Results for all evaluated rank determination methods in comparison to true rank on NewsGroup4000 and Web of Science text datasets.}
\begin{tabular}{|c|c|c|}
\hline
Method & NewsGroup4000 & Web of Science \\
\hline
MCI-RSIC & $\{4,~12,~24,~46\}$ & $\{3,~7, \text{11?},~20,~39\}$\\
Elbow & $9$ & $6$ \\
Cophenetic & $5$ & $3$ \\
Dispersion & $12$ & $3$ \\
Permutation & Und. & $64+$ \\
ARI & $3$ & $3$\\
KS-CV & N/A & N/A \\
CV2K & N/A & N/A \\
MADImput & N/A & N/A \\
\hline \hline
True Rank & $20$ & $\{35, 7\}$\\
\hline
\end{tabular}
\end{table}

\section{Discussion \& Conclusion}\label{sec:Conclusion}

In this paper, we introduced RSIC, a novel method for determining ranks of interest in NMF.
Unlike traditional methods which aim to identify a single optimal rank—often requiring extensive parameter tuning and domain-specific knowledge—our approach identifies multiple, possibly relevant ranks by analyzing the sensitivity of the reconstruction residual to different initial conditions (random initializations in the case of this paper).
This allows for a more nuanced understanding of the data's underlying structure and provides some flexibility in exploratory data analysis.

Our method identifies ``islands of stability'', which are ranks where the NMF solutions are less sensitive to initialization and, therefore, more likely to represent meaningful decompositions of the data.
We quantified this stability using the Mean Coordinatewise Interquartile Range (MCI) of the relative reconstruction error across multiple initializations.
By doing so, we highlighted ranks that consistently produce stable and interpretable factors, providing insights that single-rank methods may overlook.

We evaluated RSIC on a diverse set of datasets across various domains, including single-cell gene expression data, image datasets, and text corpora.
Our experiments demonstrated that RSIC effectively identifies ranks of interest that are consistent or close to the true underlying ranks of the data.
Of note, our method performed well on large-scale datasets where other methods tended to undershoot or where cross-validation-based approaches were infeasible due to their high computational complexity.

Comparative analysis with existing methods, including consensus-matrix methods like cophenetic correlation coefficient and dispersion coefficient, self-comparison methods like the adjusted Rand index, and cross-validation approaches, showed that RSIC is competitive and often superior in identifying meaningful ranks.

However, our method is not without limitations.
In datasets where the underlying structure is less pronounced or when the data does not exhibit clear stability islands, RSIC may suggest multiple ranks, requiring further analysis to select the most appropriate one.
For example, in the ALL-AML dataset, RSIC showed perfect stability for all ranks at or below 5, but we select 5 based on our selection criteria—this is a limitation of our method.
Additionally, while our approach reduces the computational burden compared to some methods, it still requires multiple NMF computations across a range of ranks and initializations.

For future work, we plan to refine the RSIC metric to better handle datasets with subtle or hierarchical structures.
Incorporating additional criteria, such as sparsity constraints or domain-specific knowledge, may help to further refine the ranks suggested by RSIC.
Additionally, we believe that RSIC could benefit from other types of initialization schemes, such as those based on clustering, dimensionality reduction techniques, or other randomization schemes, to further explore the space of possible initializations.
Further, the method could benefit from a window-based smoothing (e.g., Savitzky-Golay) of the MCI values to reduce the noise in the output, which could be particularly useful in datasets with high variability in the reconstruction error.
We also aim to explore the theoretical underpinnings of the observed islands of stability to provide deeper insights into why certain ranks yield more stable decompositions.

RSIC offers a robust and significantly more scalable approach for rank suggestion in NMF, taking steps toward bridging the gap between the need for meaningful data decompositions and the practical constraints of computational resources.
By providing a selection of relevant ranks and highlighting areas of stability, our method empowers practitioners to make more informed decision in exploratory data analysis and dimensionality reduction tasks.

\subsubsection*{Broader Impact Statement}
\doubleblind{The proposed RSIC method enhances the applicability of NMF in various fields by provide a more reliable and computationally efficient means of determining relevant ranks. This can benefit areas like bioinformatics, image processing, and text mining, where understanding the underlying data structure is crucial.
However, as with any data analysis tool, care must be taken into considering when drawing interpretations drawn from NMF decompositions are valid and do not inadvertently reinforce biases present in the data or researchers.}

\subsubsection*{Author Contributions}

\doubleblind{Both \textbf{Marc Tunnell} and \textbf{Erin Carrier} jointly contributed to the conceptualization of the project and different methods used in the rank suggestion method and both were involved in conceptualization of the experiments and analysis of the results. \textbf{Marc Tunnell} implemented the method, implemented the experiments, and drafted the manuscripts. \textbf{Zachary J. DeBruine} provided insight into NMF and cross-validation and was crucial in funding acquisition. \textbf{Erin Carrier} provided guidance on the methodological approach, oversaw the project, edited the paper for clarity and coherence.}

\subsubsection*{Acknowledgments}
\doubleblind{This project was in part funded by the Chan Zuckerberg Initiative Data Insights Grant DAF2022-249404. This research made use of computational resources in the Distributed Execution Network (DEN) lab at GVSU.}

\bibliography{main}
\bibliographystyle{tmlr}

\end{document}

%% file: math_commands.tex
\usepackage{amsmath,amsfonts,bm,amssymb}

\def\eqref#1{equation~\ref{#1}}

\def\1{\bm{1}}

\def\rmH{{\mathbf{H}}}

\def\rmW{{\mathbf{W}}}

\def\mA{{\bm{A}}}
\def\mB{{\bm{B}}}

\def\mG{{\bm{G}}}
\def\mH{{\bm{H}}}

\def\mM{{\bm{M}}}

\def\mR{{\bm{R}}}

\def\mW{{\bm{W}}}

\DeclareMathAlphabet{\mathsfit}{\encodingdefault}{\sfdefault}{m}{sl}
\SetMathAlphabet{\mathsfit}{bold}{\encodingdefault}{\sfdefault}{bx}{n}

\def\sS{{\mathbb{S}}}

\newcommand{\normlone}{L^1}
